# RIGA at SemEval-2016 Task 8: Impact of Smatch Extensions and Character-Level Neural Translation on AMR Parsing Accuracy

**Guntis Barzdins, Didzis Gosko**
University of Latvia, IMCS and LETA
Rainis Blvd. 29, Riga, LV-1459, Latvia
`guntis.barzdins@lu.lv, didzis.gosko@leta.lv`

## Abstract

Two extensions to the AMR smatch scoring script are presented. The first extension combines the smatch scoring script with the C6.0 rule-based classifier to produce a human-readable report on the error patterns frequency observed in the scored AMR graphs. This first extension results in 4% gain over the state-of-art CAMR baseline parser by adding to it a manually crafted wrapper fixing the identified CAMR parser errors. The second extension combines a per-sentence smatch with an ensemble method for selecting the best AMR graph among the set of AMR graphs for the same sentence. This second modification automatically yields further 0.4% gain when applied to outputs of two nondeterministic AMR parsers: a CAMR+wrapper parser and a novel character-level neural translation AMR parser. For AMR parsing task the character-level neural translation attains surprising 7% gain over the carefully optimized word-level neural translation. Overall, we achieve smatch F1=62% on the SemEval-2016 official scoring set and F1=67% on the LDC2015E86 test set.

## 1 Introduction

Abstract Meaning Representation (AMR) (Banarescu et al., 2013) initially was envisioned as an intermediate representation for semantic machine translation, but has found applications in other NLP fields such as information extraction.

For SemEval-2016 Task 8 on Meaning Representation Parsing we took a dual approach: besides developing our own neural AMR parser, we also extended the AMR smatch scoring tool (Cai and Knight, 2013) with a rule-based C6.0 classifier[1] to guide development of an accuracy-increasing wrapper for the state-of-art AMR parser CAMR (Wang et al., 2015a; 2015b). A minor gain was also achieved by combining these two approaches in an ensemble.

The paper starts with the description of our smatch extensions, followed by the description of our AMR parser and wrapper, and concludes with the results section evaluating the contributions of described techniques to our final SemEval result.

## 2 Smatch Extensions

We describe two extensions[2] to the original AMR smatch scoring script. These extensions do not change the smatch algorithm or scores produced, but they extract additional statistical information helpful for improving results of any AMR parser, as will be illustrated in Section 3.

### 2.1 Visual Smatch with C6.0 Classifier Rules

The original AMR smatch scoring metric produces as output only three numbers: precision, recall and F1. When developing an AMR parser, these three numbers alone do not reveal the actual mistakes in the AMR parser output (we call it *silver* AMR) when compared to the human-annotated *gold* AMR.

---

[1] Available at http://c60.ailab.lv

[2] Available at https://github.com/didzis/smatchTools

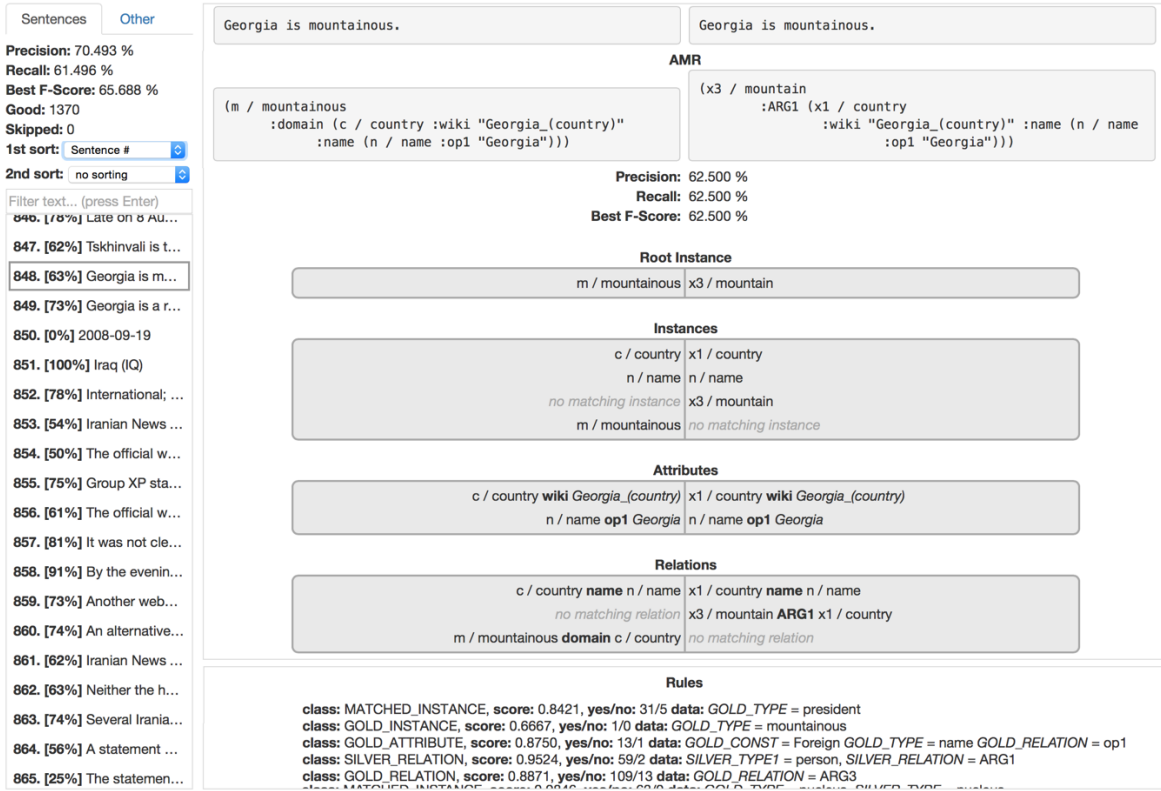

**Figure 1.** Visual smatch with Rules. Left pane shows the document content and statistics. Right pane shows single sentence gold AMR (left) and silver AMR (right) along with smatch aligned instance, attribute, relation AMR graph edges. The bottom pane shows C6.0 classifier generated rules describing the common error patterns found in the document.

The first step in alleviating this problem is visualizing the mappings produced by the smatch algorithm as part of the scoring process. Figure 1 shows such smatch alignment visualization where gold and silver AMR graphs are first split into the edges, which are further aligned through variable mapping. The smatch metric measures success of such alignment – perfect alignment results in F1 score 100% while incomplete alignment produces lower scores.

The visualization in Figure 1 is good for manual inspection of incomplete AMR alignments in individual sentences. But it still is only marginally helpful for AMR parser debugging, because the data-driven parsers are expected to make occasional mistakes due to the training data incompleteness rather than due to a bug in the parser.

Telling apart the repetitive parser bugs from the occasional training data incompleteness induced errors is not easy and to invoke the required statistical mechanisms we resorted to a rule-based C6.0 classifier (Barzdins et al., 2014; 2015), a modification of the legacy C4.5 classifier (Quinlan, 1993). The classifier is asked to find most common patterns (rules) leading to some AMR graph edges to appear mostly in the gold, silver, or matched class after the smatch alignment. The bottom part of Figure 1 illustrates few such rules found by C6.0. For example, the second rule relates to the visualized sentence and should be read as "if the instance has type *mountainous*, then it appears 1 time in the gold graphs and 0 times in the silver graphs of the entire document". Similarly the third rule should be read as "word *Foreign* appears 13 times as :op1 of name in the gold graphs, but only 1 time in the silver graphs of the entire document" – such 13 to 1 ratio likely points to some capitalization error in the parser pipeline. The generated rules can be sorted by their statistical impact score calculated as Laplace ratio $(p+1)/(p+n+2)$ from the number of correct $p$ and wrong $n$ predictions made by this rule.

Classifier generated rules were the key instrument we used to create a bug-fixing wrapper for the CAMR parser, described in Section 3.1. We fixed only bugs triggering error-indicating-rules with the impact scores above 0.8, because Laplace ratio strongly correlates with the smatch score impact of the particular error.

## 2.2 Smatch Extension for Ensemble Voting

The original smatch algorithm is designed to compare only two AMR documents. Meanwhile CAMR parser is slightly non-deterministic in the sense that it produces different AMRs for the same test sentence, if trained repeatedly. Randomly choosing one of the generated AMRs is a suboptimal strategy. A better strategy is to use an ensemble voting inspired approach: among all AMRs generated for the given test sentence, choose the AMR which achieves the highest average pairwise smatch score with all the other AMRs generated for the same test sentence. Intuitively it means that among the non-deterministic options we choose the "prevalent" AMR.

Multiple AMRs for the same test sentence can be generated also from different AMR parsers with substantially different average smatch accuracy. In this case all AMRs still can participate in the scoring, but weights need to be assigned to ensure that only AMRs from the high-accuracy parser may win.

## 3 AMR Parsers

We applied the smatch extensions described in the previous Section to two very different AMR parsers.

### 3.1 CAMR Parser with Wrapper

We applied the debugging techniques from Section 2.1 to the best available open-source AMR parser CAMR[3]. The identified bug-fixes were almost entirely implemented as a CAMR parser wrapper[4], that runs extra pre-processing (normalization) step on input data and extra post-processing step on output data. Only minor modifications to CAMR code itself were made[5] to improve the performance on multi-core systems and to fix date normalization problems.

Our CAMR wrapper tries to normalize the input data to the format recognized well by CAMR and to fix some systematic discrepancies of annotation style between the actual CAMR output and the expected gold AMRs. The overall gain from our wrapper is about 4%.

The following normalization actions are taken during pre-processing step, together accounting for about 2% gain:

1. number normalization from a lexical (e.g. "seventy-eight"), semi-lexical (e.g. "5 million") or multi-token digital (e.g. "100,000" or "100 000") format to a single token digital format (e.g. "100000");

2. currency normalization from a number (any format mentioned in previous step) together with a currency symbol (e.g. "$ 100") to a single token digital number with the lexical currency name (e.g. "100 dollars");

3. date normalization from any number and lexical mix to an explicit eight-digit dash separated format "yyyy-mm-dd".

Small modifications had to be made to the baseline JAMR (Flanigan et al., 2014) aligner used by CAMR to reliably recognize the "yyyy-mm-dd" date format and to correctly align the date tokens to the graph entries (by default JAMR uses "yymmdd" date format that is ambiguous regarding century and furthermore can be misinterpreted as a number).

The rules for date normalization were extracted from the training set semi-automatically using C6.0 classifier by mapping date-entities in the gold AMR graphs and corresponding fragments in input sentences.

Additionally, all wiki edges were removed from the AMR graphs prior to training, because CAMR does not handle them well; this step ensures that CAMR is trained without wiki edges and therefore will not insert any wiki entries in the generated AMR. Instead, we insert wiki links deterministically during the post-processing step.

During post-processing step the following modifications are applied to the CAMR parser generated AMR graphs, together accounting for about 2% gain:

1. nationalities are normalized (e.g. "Italian" to "Italy");

2. some redundant graph leafs not carrying any semantic value are removed (e.g. "null-edge");

3. wiki links are inserted deterministically next to "name" edges using gazetteer extracted from the training data and extended with the complete list of countries and nationalities (wiki value is selected

---
[3] https://github.com/Juicechuan/AMRParsing
[4] https://github.com/didzis/CAMR/tree/wrapper
[5] Modified CAMR at https://github.com/didzis/CAMR

based on the parent concept and content of the "name" instance);

About 1% additional gain comes from the observation that CAMR parser suffers from overfitting: it achieves optimal results when trained for only 2 iterations and with empty validation set.

### 3.2 Neural AMR Parser

For neural translation based AMR parsing we used simplified AMRs without wiki links and variables. Prior to deleting variables, AMR graphs were converted into trees by duplicating the co-referring nodes. Such AMR simplification turned out to be nearly loss-less, as a simple co-reference resolving script restores the variables with average F1=0.98 smatch accuracy.

We trained a modified TensorFlow seq2seq neural translation model[6] with attention (Abadi et al., 2015; Sutskerev et al., 2014; Bahdanau et al., 2015) to "translate" plain English sentences into simplified AMRs. For the test sentence in Figure 1 it gives following parsing result:

```
(mountain-01
   :ARG1 (country
      :name (name :op1 "Georgia"))
```

Apart from a missing bracket this is a valid (although slightly incorrect) simplified AMR. Note that this translation has been learned in "closed task" and "end-to-end" manner only from the provided AMR training set without any external knowledge source. This explains overall lower accuracy of the neural AMR parser compared to CAMR, which uses external knowledge from wide coverage parsing models of BLLIP[7] parser (Charniak and Johnson, 2005). The neural AMR parser accuracy is close to CAMR accuracy for short sentences up to 100 characters, but degrades considerably for longer sentences.

We optimized TensorFlow seq2seq model hyperparameters within the constraints of the available GPU memory: 1 layer or 400 neurons, single bucket of size 480, each input and output character tokenized as a single "word", vocabulary size 120 (number of distinct characters), batch size 4, trained for 30 epochs (4 days on TitanX GPU).

Operating seq2seq model on the character-level (Karpathy, 2015; Chung et al., 2016; Luong et al., 2016) rather than standard word-level improved smatch F1 by notable 7%. Follow-up tests (Barzdins et al., 2016) revealed that character-level translation with attention improves results only if the output is a syntactic variation of the input (as is the case for AMR parsing), but for e.g. English-Latvian translation gives inferior results due to attention mechanism failing to establish character-level mappings between the English and Latvian words.

| Parser | F1 on LDC2015E86 test set | F1 on the official eval set |
|---|---|---|
| JAMR (baseline) | 0.576 | |
| CAMR (baseline) | 0.617 | |
| CAMR (no valid.set, 2 iter.) | 0.630 | |
| Neural AMR (word-level) | 0.365 | |
| Neural AMR (char-level) | 0.433 | 0.376 |
| CAMR+ wrapper | 0.667 | 0.616 |
| Ensemble of CAMR+ wrapper and NeuralAMR (char-level) | 0.672 | **0.620** |

**Table 1:** Smatch scores.

## 4 Results

Table 1 shows smatch scores for various combinations of parsers and thus quantifies the contributions of all methods described in this paper. We improved upon CAMR rather than JAMR parser due to better baseline performance of CAMR, likely due to its reliance on the wide coverage BLLIP parser.

The CAMR parser wrapper (result of Sections 2.1 and 3.1) is the largest contributor to our results. The weighed ensemble of 3 runs of CAMR+wrapper and 1 run of neural AMR parser (Sections 2.2 and 3.2) gave an additional boost to the results. Including the neural AMR parser in the ensemble doubled the gain – apparently it acted as an efficient tiebreaker between the similar CAMR+wrapper outputs.

## 5 Conclusions

Although our results are based on CAMR parser, the described debugging and ensemble approaches are likely applicable also to other AMR parsers.

## Acknowledgments

This work was supported in part by the Latvian National research program SOPHIS under grant agreement Nr.10-4/VPP-4/11 and in part by H2020 SUMMA project under grant agreement 688139/H2020-ICT-2015.

---

[6] https://github.com/didzis/tensorflowAMR

[7] https://github.com/BLLIP/bllip-parser